\documentclass{INTERSPEECH2023}

\usepackage{xcolor}

\interspeechcameraready

\title{Using Text Injection to Improve Recognition of Personal Identifiers in Speech}
\name{Yochai Blau, Rohan Agrawal, Lior Madmony, Gary Wang, Andrew Rosenberg, Zhehuai Chen, Zorik Gekhman, Genady Beryozkin, Parisa Haghani, Bhuvana Ramabhadran}
\address{
  Google}
\email{\{yochaib,rohanag,liomad,wgary,rosenberg,zhehuai,zorik,genady,parisah,bhuv\}@google.com}

\begin{document}

\maketitle
 
\begin{abstract}

Accurate recognition of specific categories, such as persons' names, dates or other identifiers is critical in many Automatic Speech Recognition (ASR) applications. 
As these categories represent personal information, ethical use of this data including collection, transcription, training and evaluation demands special care.  One way of ensuring the security and privacy of individuals is to redact or eliminate Personally Identifiable Information (PII) from collection altogether.  However, this results in ASR models that tend to have lower recognition accuracy of these categories. 
We use text-injection to improve the recognition of PII categories by including fake textual substitutes of PII categories in the training data using a text injection method. We demonstrate substantial improvement to Recall of Names and Dates in medical notes while improving overall WER.  For alphanumeric digit sequences we show improvements to Character Error Rate and Sentence Accuracy.

\end{abstract}

\section{Introduction}

While a lot of speech is publicly broadcast or shared online to a vast audience, speech recognition applications frequently interact with more private communications such as dictation, call-center conversations or conversing with a digital assistant.  Automatic speech recognition (ASR) systems are trained using transcribed speech, and operate best when their training data matches the context in which they are used.  This raises a question of how to train ASR to work well on private communications, while being sensitive to the complexities of collecting and transcribing training material from private contexts.

We engage with this question in the context of healthcare. Accurate transcription of medical speech is key to a growing set of applications, including dictation of clinical notes and voice assistance. 
However, ethical collection and transcription of training data for medical ASR requires care.  Medical data is exceptionally sensitive and respecting patients' privacy in any collection effort is essential. De-identification, a process which includes the redaction of PII tokens, is a common practice to protect users' private data. In the process, the audio segment is replaced by silence and the text is either removed or replaced by a special markup tag. De-identification is used extensively for healthcare data and is often required by regulations such as the US HIPAA Act \cite{hipaa}. 

In contexts such as call-center applications, speech utterances are shorter and may be in response to prompts like ``Please confirm your date of birth'' or ``Please say your user id''.  In this case, de-identification results in completely eliminating the utterance.  
In both domains, using such de-identified datasets for training ASR models makes recognition of these classes of PII terms operate at a higher error rate than the surrounding speech.

Prior work has focused around replacing PII in speech with arbitrary synthesized speech from the same category \cite{flechl22_interspeech} (for example, named entities, number sequences) or ensuring that these PII tokens/sequences are adequately covered by a language model, as done in \cite{parity_2017}.
Text injection has recently emerged as an effective way to leverage text-only data for ASR training in a single model \cite{chen2022maestro,bapna2022mslam,sainath2023joist,thomas2022textogram} without an LM. Techniques such as the one described in ~\cite{chen2022maestro} have shown promise for zero-transcribed speech adaptation. In ~\cite{wang2023understanding}, Wang \& Kastner et al.~demonstrated that text-injection can effectively address domain transfer on diverse SpeechStew corpora \cite{chan2021speechstew} with little to no in-domain speech. 
Motivated by this, we propose to generate a text dataset that contains fake terms instead of the redacted PII and use text-injection during model training. This eliminates the need for any sensitive transcribed training data and allows effective use of de-identified medical datasets.

This paper addresses three specific data challenges when training a model for private contexts with the use of text-injection in ASR training.
\begin{itemize}
    \item Domain adaptation: Transcribed speech from private contexts is much harder to come by than public contexts. We show that text-injection along with a small amount of transcribed speech can effectively perform domain adaptation to Medical ASR resulting in a 11.5 reduction to WER and improvements of 1\% to Names and 11\% to Dates  (Section \ref{ssec:redacted}). 
    \item Redaction: Available in-domain data has redacted PII from the transcript with corresponding speech replaced by silence. To mitigate this, we train on generated, spoken text similar to the PII, which allows for substantial improvement to sensitive term recognition without using any individual's protected data; compared to training on redacted speech data, Names and Dates recall is improved by 8\% and 13\% (Section \ref{ssec:redacted}).
    \item Elimination:  De-identification may eliminate utterances that contain sensitive terms. For example, short utterances containing alphanumeric and digit sequences are eliminated.  We show that text-injection can substantially improve recognition of these classes of potentially sensitive information with a full-sequence Sentence Accuracy improvement of 3.2\% (and CER improvement of 1.4\%) (Section \ref{ssec:personal-identifiers}).

\end{itemize}

\section{Related Work}

The complexity of recognizing personal identifiers in medical speech has recently been studied in \cite{flechl22_interspeech}. This work shows that PII recognition accuracy declines when training an end-to-end model on de-identified data. The authors propose to generate artificial audio/text pairs with synthetic identifiers, and show that training on such data restores most of the performance degradation. Artificial audio/text pairs are generated in a two-step process: first, PII in the text is replaced with random data. Then, corresponding audio is generated or spliced with matching fragments. Generating or manipulating the audio is more demanding then replacing identifiers in the text 
and also requires accurate word-level timestamps. In our work we take this concept one step further, by improving PII recognition with text replacements alone, and \emph{without} any audio generation or manipulation.
Other work has studied medical domain ASR \cite{parity_2017,chiu18_interspeech,mani,zhou_2018}, yet these have not reported findings on PII recognition accuracy and the effect of training on de-identified data.

Regarding text-injection in ASR training, this work builds on the architecture described in \cite{chen2022maestro}.  This is described in detail in Section \ref{sec:details}.  However, there is a set of related approaches and architectures that similarly use separate speech and text encoders that feed into a shared encoder \cite{bapna2022mslam,sainath2023joist,thomas2022textogram}.  Additionally, a similar thread of work trains on speech and text data operates by first converting the text data to speech via TTS \cite{rosenberg2019synthesized,chen2022tts4pretrain,fazel2021synthasr,zheng2021using,wang2020improving}.  These approaches, in effect, use speech as an intermediate representation to perform text injection.  

\section{Text-Injection and ASR Details}
\label{sec:details}
The text-injection model used in this work includes a speech encoder, a text encoder with a learned duration model, a shared encoder, decoder and an alignment decoder following~\cite{chen2022maestro}. The speech encoder consists of 4 causal conformer layers. The shared encoder consists of 3 causal conformer layers and 10 non-causal conformer layers, each with model dimension of 512. The text encoder contains 2 conformer layers and 4 lightweight convolutional upsampling layers \cite{elias2021parallel}. HAT decoders \cite{variani2020hat} with $v^2$ embeddings \cite{ghodsi2020stateless} are used in both decoder and an alignment decoder with the distinction of the former produces word-piece models as text outputs and the latter uses phonemes as model units to get speech-text alignments. The overall model contains 165M parameters, with an additional 58M in the text encoder which is only used during training.

Text-injection training involves speech-text and text-only training paths in a curriculum fashion. Initially, speech-text training is used to minimize a consistency loss, a HAT decoder loss and an alignment decoder loss, for the duration model. Consistency loss ensures that we learn corresponding mappings from text to speech embedding. The duration model is used to up-sample text embedding before being fed to the shared encoder. After beginning training with speech-text loss, we enable text-only loss to be able to perform text-injection training. The text-only training step involves minimizing an aligned masked-language model loss. Further details of text injection architecture and training can be found in ~\cite{chen2022maestro}.

In all experiments, we train a text-injection model on supervised only data for 10k steps first so that the speech and text encoders generate consistent features. We continue training with target domain text or targeted terms as described in the following sections.  While training with text alone is feasible after the initial supervised training, we find that including supervised data during training helps to stabilize the model behavior.

\section{Data Description}

\ifinterspeechfinal
We are aware of the sensitive nature of speech recognition research particularly on personal identifiers. Therefore, we
ensure that this work abides by the Google AI Principles \cite{ai-principles}.
\fi

\subsection{Medical domain datasets}
The \textit{Medical Audio} dataset consists of dictations of clinical notes by healthcare professionals from a variety of medical specialties. During the transcription process any potentially identifying information was redacted. The corresponding audio segment was replaced by silence and a special markup was placed in the transcript to indicate the type of the data that was removed. Examples of the markup tags include \texttt{PATIENT\_NAME}, \texttt{MEDICAL\_PROFESSIONAL\_NAME}, \texttt{AGE} and \texttt{DATE}.
The resulting dataset was certified to comply with de-identification requirements of the US HIPAA Privacy Rule \cite{hipaa}.

The \textit{Medical Text} dataset consists of the transcriptions from the \textit{Medical Audio} dataset (without the audio recordings). As mentioned above, these transcriptions include markup tags for PIIs such as names, addresses, dates, etc.~that were redacted. In this dataset, we replace four types of markup tags (names, identifying numbers, dates and ages) with fake random data. For example, any name tag is replaced by a random name from a real-world distribution, each redacted digit is replaced with a random digit between 0-9, and so on. Note that replacing redacted PII information in text is quite easy and straightforward. However, splicing synthetic speech within speech recordings where the PII was silenced in a natural manner is challenging and error-prone, especially in the case of fast-paced speech recorded in a noisy environment as it is in the \textit{Medical Audio} dataset.

The \textit{Synthetic Notes} dataset is a collection of fake dictations, recorded by clinicians in a noise-free environment and consists of 14.5 hours of audio. The dataset consists of synthetic hospital visit summaries which include fake identifiers such as names, IDs, dates, etc. The dataset creation process started with creation of fake hospital visit summaries by trained clinicians. Later, a separate group of clinicians simulated dictations of hospital notes based on these visit summaries, which were then independently transcribed.

The \textit{Synthetic Names} dataset is a 14-hours long dataset that consists of short phrases (average length 7.6s) that resemble phrases in the dictation of clinical notes and include references to patient or clinician names. First, medical specialists created 550 textual templates with placeholders for names, ensuring diversity of name formats. Each template was recorded in a quiet environment by 17 different speakers, with placeholders substituted by names, each time randomly drawn from a real-world distribution. All name occurrences are tagged, which allows to compute a recall metric for names alone (along with more global metrics such as WER).

\begin{table}[!htp]\centering
\caption{Dataset descriptions.}\label{tab:datasets}
\scriptsize
\begin{tabular}{|p{0.8cm}|p{0.7cm}|p{0.8cm}|p{3.7cm}|}\hline
\textbf{Name} &\textbf{Size (hours)} &\textbf{Type} &\textbf{Description} \\\hline
\multicolumn{4}{| c |}{Training data} \\\hline
Captions &377K &Speech & Speech from videos and corresponding captions. \\\hline
Medical Audio &5K &Speech & Clinicians dictating notes. All private information is redacted. \\\hline
Medical Text  &- &Text & The transcriptions from the \textit{Medical Audio} dataset, where redacted identifiers are replaced with random data. \\\hline
\multicolumn{4}{| c |}{Testing data} \\\hline
Synthetic Notes &14.5 &Speech & Synthetic hospital notes dictated by clinicians, including (fake) identifiers. \\\hline
Synthetic Names &14 &Speech & Synthetic short phrases from clinical notes, including (fake) names (patient/clinician/family member/etc.). \\
\hline
\end{tabular}
\end{table}

\vspace{-5mm}
\subsection{Alphanumeric Sequence Identifier Datasets}
\label{ssec:personal-identifier}

For personal identifier text-injection train data, we use 100M sequences each of alphanumeric sequences and text sequences. The length of these sequences is sampled from a skewed normal distribution $X (\mu=10,\,\sigma=5) + 3 $. For digit sequences, characters from {0-9} are sampled uniformly. For alphanumeric sequences, characters from the alphanumeric set {0-9,a-z} are sampled uniformly. 10\% of sequences are chosen at random, and character repeats of length 2, 3 or 4 are added. This is done because character or digit repetitions are a challenging use case for an ASR system.

The \textit{spoken digit sequence} test set consists of 2146 utterances (average length 4.3s) with randomly generated digit sequences, spoken by voice actors. 

The \textit{spoken alphanumeric sequence} test set consists of 1,992 utterances (average length 5.2s) with randomly generated alphanumeric sequences, spoken by voice actors. The voice actors in both sequence test sets have acted out hesitations and pauses while speaking these sequences.

The following TTS datasets contain utterances synthesized by a commercial American English TTS system sampled from 6 voices. In this work, we use the TTS datasets as test data, and the spoken sequence data as either test or training material (Section \ref{ssec:personal-identifiers}).

\textit{TTS digit sequence} test set consists of 1991 utterances (average length 5.4s) with randomly generated digit sequences.

The \textit{TTS alphanumeric sequence} test set consists of 2000 utterances (average length 5.7s) with randomly generated alphanumeric sequences.

The \textit{TTS digit sequence repetition} consists of 3188 utterances (average length 5.4s) with randomly generated digit sequences which have at minimum, 1 consecutive digit repetition. 1001 utterances have 2, 3 and 4 consecutive repetitions and 185 utterances have more than 4 repetitions.

The \textit{TTS alphanumeric sequence repetition} consists of 2179 utterances (average length 5.7s) with randomly generated alphanumeric sequences which have at minimum, 1 consecutive character repetition. 1001 utterances have 2, 3 consecutive repetitions and 177 utterances have more than 4 repetitions. 

\subsection{General datasets}
\label{sec:general-data}

The base model (B1) is trained with close to 300k hours of multidomain utterances including YouTube, Telephony and Dictation for US English as described in \cite{narayanan2019recognizing}. Multi-condition training \cite{kim2017mtr}, random 8kHz down-sampling \cite{Li12} are applied on the training data.   Importantly, any utterance that includes alphanumeric and digit sequences are eliminated from transcribed training data, as they may represent personal identifiers.   This data is only included in initial checkpoint training, described as A1/B1 in Section \ref{sec:eliminated}, and is not further used during finetuning.

Along with the medical domain data described above, we also include 377k hours of YouTube {\it Captions} data as described in \cite{liao2013confisland} to maintain strong general purpose ASR performance. 
While this material may incidentally include medical content, no particular targeting, selection or filtering for in-domain material was performed.

\section{Recognizing Redacted and Eliminated Terms}
\label{sec:eliminated}

\subsection{Identifiers in medical speech}
\label{ssec:redacted}

All models used the architecture described in Section \ref{sec:details}, and use a base checkpoint, B1, trained on multidomain data (Section \ref{sec:general-data}).  The  A1  model was trained on the (non-medical) \textit{Captions} dataset, in which medical terms are relatively uncommon.  This shows clear domain-transfer effects; the WER on the medical-term rich \textit{Synthetic Notes} dataset is high, with a WER of 14.6\% (Table \ref{tab:medicalexps}).

Adapting this model to the medical domain is done by training on a mixture of \textit{Captions} and \textit{Medical Audio} datasets with a 90\%/10\% ratio, resulting in model A2. The \textit{Synthetic Notes} WER drops significantly, and we observe higher recall for medical entities such as conditions and medications. Yet, for other entities such as names and dates 
the recall metrics degrade. This is a direct consequence of names, dates and other identifiers being redacted from the  \textit{Medical Audio} dataset.

To improve error rates on PII, model T1 is trained on the same mixture of speech data as for A2, with the \textit{Medical Text} dataset injected during training. This textual data contains identifiers such as names and dates, leading to a 8\%/13\% boost in names/dates recall, respectively. In Table \ref{tab:examples} we show two cases were model T1, which has been exposed to medical text with (fake) identifiers, performs better in transcribing examples from the \textit{Synthetic Names} dataset.  We also observe that the introduction of medical text further improves the overall WER from 3.3 to 3.1, demonstrating the value of additional in-domain text to address domain transfer as well as targeted terms.

\begin{table}[!htp]\centering
\vspace{-2mm}
\caption{Word error rate for Synthetic Notes and medical-term/name/date recall for Synthetic Names.}\label{tab:medicalexps}
\scriptsize
\begin{tabular}{|m{0.3cm}|m{2.3cm}|m{0.9cm}|m{0.7cm}|m{0.6cm}|m{0.6cm}|}\hline
&\textbf{Datasets} &\textbf{Synthetic Notes WER} &\textbf{Medical term recall} &\textbf{Names recall} &\textbf{Dates recall} \\\hline
\textbf{B1} &Multidomain (MD)&14.6 &82.6\% &60\% &66\%\\\hline
\textbf{A1} &MD, Captions &14.7 &82.1\% &59\% &72\% \\\hline
\textbf{A2} &MD, Captions\newline Medical Audio &3.3 &97.4\% &53\% &64\% \\\hline
\textbf{T1} &MD, Captions\newline Medical Audio\newline Medical Text (injected) &3.1 &97.5\% &61\% &77\% \\
\hline
\end{tabular}
\vspace{-2mm}
\end{table}

\begin{table}[!htp]\centering
\vspace{-2mm}
\caption{Examples of name mistranscriptions from Synthetic Names corpus.  Note: All examples are fake descriptions of hypothetical scenarios. }\label{tab:examples}
\scriptsize
\begin{tabular}{|l|m{6.5cm}|}\hline
\textbf{Model} &\textbf{Transcription}\\\hline
\hline
\textbf{Truth} &Scarlett Kathleen Ibarra was admitted to the floors\\\hline
\textbf{A2} &Scarlet {\color{red}Caffeine} Ebara was admitted to the floors\\\hline
\textbf{T1} &Scarlett Kathleen Ibarra was admitted to the floors\\\hline\hline
\textbf{Truth} &Oliver Barry Matthews is adamant that the patient had no past surgeries\\\hline
\textbf{A2} &{\color{red}All of our bearing math uses} adamant that the patient had no past surgeries\\\hline
\textbf{T1} &Oliver Barry Matthews is adamant that the patient had no past surgeries\\\hline
\end{tabular}
\vspace{-2mm}
\end{table}

The training data used for models A1, A2 and T1 above, includes a large general-domain multidomain and \textit{Captions} dataset. This dataset includes identifying entities such as names and dates to some extent. To isolate the effect of PII redaction, we train a cold-start model A3 only on the \textit{Medical Audio} dataset (where all identifiers are redacted). The results of this experiment are in Table \ref{tab:medicalablation}. All error metrics degrade, but particularly noticeable is the names recall which drops to 1\%, as the model has not seen \emph{any} names during training. Yet performance can be improved by injecting text which includes fake identifiers, even without a single audio example of such entities, as seen for model T2. While the metrics indicate that text injection alone cannot match the performance of T1, the recall of Names and Dates are improved by a factor of 1800\% and 207\%, respectively.  Moreover, the introduction of medical text improves the WER performance of the model from 6.1 to 4.1.

\begin{table}[!htp]\centering
\caption{Ablation study: training only on medical data.}\label{tab:medicalablation}
\scriptsize
\begin{tabular}{|m{0.3cm}|m{2.3cm}|m{0.9cm}|m{0.7cm}|m{0.6cm}|m{0.6cm}|}\hline
&\textbf{Datasets} &\textbf{Synthetic Notes WER} &\textbf{Medical term recall} &\textbf{Names recall} &\textbf{Dates recall} \\\hline
\textbf{A3} &Medical Audio &6.1 &96.2\% &1\% &28\% \\\hline
\textbf{T2} &Medical Audio\newline Medical Text (injected) &4.1 &97.2\% &18\% &58\% \\
\hline
\end{tabular}
\vspace{-2mm}
\end{table}

\subsection{Recognizing Alphanumeric Sequence Identifiers}
\label{ssec:personal-identifiers}

The architecture of the base ASR model B1 is the same RNN-T model with a cascade encoder and HAT decoder with $v^2$ embeddings in Section \ref{ssec:redacted}.
The baseline text injection model, M1 was trained with the same paired text data as B1, and additional base text injection data (not related to target domain) consists of 100B anonymized sentences across different domains.  Note that this text injection model is trained in arbitrary text to improve core ASR performance rather than targeting any particular domain or type of lexical content.

We aim to improve recognition of alphanumeric and digit sequence identifiers. Thus, we perform a round of text injection training based on the initial model described above, specifically, the personal identifier text data (cf. Section \ref{ssec:personal-identifier}). Results of these experiments measured by character error rate (CER) and Sentence Accuracy (SACC) in Tables \ref{tab:cer-seq} and \ref{tab:sacc-seq}. Character error rate is more fine grained, by being able to measure individual errors. However, when recognizing an identifier it is crucial to recognize the full sequence; this is measured by SACC.

\begin{table}[!htp]\centering
\caption{CER on alphanumeric and digit test sets.}\label{tab:cer-seq}
\scriptsize
\begin{tabular}{|p{2cm}|p{0.6cm}|p{1cm}|p{1.1cm}|p{1.1cm}|}\hline
\textbf{Dataset} &\textbf{B1: Base model} &\textbf{M1: B1 + base text injection} &\textbf{M2: M1 + \newline in-domain text inj.} &\textbf{M3: M2+\newline Spoken datasets}\\\hline
{Spoken digit} &3.1 &2.4 &2 &-\\
{TTS digit rep.} &15.2 &15.7 &14.5 &17.0 \\
{TTS digit} &1.7 &1 &1.1 &0.8 \\\hline
{Spoken alphanum} &8.6 &7.3 &6.8 &-\\
{TTS alphanum rep.} &12.1 &11.9 &9.3 &8.4\\
{TTS alphanum} &9.2 &8.5 &8 &6.2\\\hline
{average  CER} &8.3 &7.8 &6.9 &-\\
\hline
\end{tabular}
\vspace{-2mm}
\end{table}

\begin{table}[!htp]\centering
\vspace{-2mm}
\caption{SACC on alphanumeric and digit test sets.}\label{tab:sacc-seq}
\scriptsize
\begin{tabular}{|p{2cm}|p{0.6cm}|p{1cm}|p{1.1cm}|p{1.1cm}|}\hline
\textbf{Dataset} &\textbf{B1: Base model} &\textbf{M1: B1 + base text injection} &\textbf{M2: M1 + \newline in-domain text inj.} &\textbf{M3: M2+\newline Spoken datasets}\\\hline
{Spoken digit} &93.8 &94.7 &95.2 &-\\
{TTS digit rep.} &67.4 &69.1 &70.4 &64.3\\
{TTS digit} &96.8 &96.6 &97.1 &96.8\\\hline
{Spoken alphanum} &73.6 &75.5 &76.5 &-\\
{TTS alphanum rep.} &54.2 &53.5 &63.9 &63.6\\
{TTS alphanum} &66.0 &67.4 &68.1 &72.4\\\hline
{average SACC} &75.3 &76.1 &78.5 &-\\
\hline
\end{tabular}
\end{table}

We use CER as a metric instead of word error rate for evaluation of personal identifier recognition because CER is a better fit for these sequences; an utterance may contain a single ``word'' like {\it A1B2C3}. Normalization helps to ignore formatting issues with respect to such sequences. As can be seen from Table \ref{tab:cer-seq}, even the addition of base text injection data which is not related to target domain (M1) improves CER from 8.3 to 7.8, a reduction of 6\% from B1. Additional training on domain specific text injection data (M2), further reduces normalized CER to 6.9, a reduction of 14.1\% from M1. Correspondingly when considering SACC, after adding base text injection data (M1), we see an increase from 75.3 to 76.1, a modest increase of 1.1\% from B1. SACC further increases to 78.5 after training on domain specific text injection data (M2), an increase of 3.1\% from M1. 
In M3, we add Spoken sequence data (4138 utterances) to our training corpus. The motivation for this experiment is to show the effect of adding a small amount of paired speech-text data in addition to text-injection training can further help. This further reduces CER on alphanumeric test sets, but gives mixed results in digit test sets. 

Catastrophic forgetting is a concern when fine-tuning a pretrained model on material from a narrow domain, as we are doing here with text-injection of sequences.  To assess the impact of this we assess performance on a Short Utterance test set representative of spoken search queries, and TTS test sets focusing on rare proper names from 5 different domains  (Table \ref{tab:forgetting}).  While we see a minor regression between M1 and M2, the performance is still substantially better than B1, the base model with no text-injection.  This suggests that this regression can be mitigated with tuning of a ratio and schedule of the text-injection training.

\begin{table}[!htp]\centering
\vspace{-4mm}
\caption{WER on broad domain test sets.}\label{tab:forgetting}
\scriptsize
\begin{tabular}{|p{2cm}|p{1cm}|p{1cm}|p{1.3cm}|}
\hline
\textbf{Dataset} &\textbf{B1: Base model} &\textbf{M1: B1 + base text injection} &\textbf{M2: M1 + \newline in-domain text inj.} \\\hline
{Short Utterances} &6.7 &5.6 &5.7 \\
{TTS rare terms 1} &14.5 &11.3 &11.9 \\
{TTS rare terms 2} &9.8 &8.6 &8.9 \\
{TTS rare terms 3} &37.9 &33.5 &34.4 \\
{TTS rare terms 4} &21.9 &17.9 &18.8 \\
{TTS rare terms 5} &24.9 &20.7 &21.4 \\
\hline
\end{tabular}
\vspace{-4mm}
\end{table}

\section{Conclusions}

Due to the sensitive nature of medical speech and personal identifiers, substantial care must be taken in the collection, transcription and use of this data, often requiring the removal of personal identifiers from the data. Models trained on such data have higher error rates when recognizing such identifiers. In this paper, we demonstrate that text-injection is an effective approach to improving ASR performance of models trained on de-identified speech, including improved recognition of personal identifiers. Text-injection can leverage text where speech is unavailable, and synthetic text is simple to construct, frequently indistinguishable from real transcripts, and not associated with any person's private communication. 
In Medical ASR where names and dates are redacted from speech and transcripts, we find core WER improves from 3.3 to 3.1, and recall of names and dates are improved by 8\% and 13\% respectively. For short utterances containing alphanumeric and digit sequences which can be used as personal identifiers in a broad range of contexts, we improve SACC by an average of 3.2\% while CER is reduced from 8.3\%  to 6.9\%

\bibliographystyle{IEEEtran}
\bibliography{arxiv}

\end{document}